# Surface registration using genetic algorithm in reduced search space

Vedran Hrgetić, Tomislav Pribanić

Department of Electronic Systems and Information Processing
Faculty of Electrical Engineering and Computing
Zagreb, Croatia
hrgetic.vedran@gmail.com,
tomislav.pribanic@fer.hr

**Abstract** – Surface registration is a technique that is used in various areas such as object recognition and 3D model reconstruction. Problem of surface registration can be analyzed as an optimization problem of seeking a rigid motion between two different views. Genetic algorithms can be used for solving this optimization problem, both for obtaining the robust parameter estimation and for its fine-tuning. The main drawback of genetic algorithms is that they are time consuming which makes them unsuitable for online applications. Modern acquisition systems enable the implementation of the solutions that would immediately give the information on the rotational angles between the different views, thus reducing the dimension of the optimization problem. The paper gives an analysis of the genetic algorithm implemented in the conditions when the rotation matrix is known and a comparison of these results with results when this information is not available.

## I. Introduction

Surface registration is an important step in the process of reconstructing the complete 3D object. Acquisition systems can mostly give only a partial view of the object and for its complete reconstruction should capture from the multiple views. The task of the surface registration algorithms is to determine the corresponding surface parts in the pair of observed clouds of 3D points, and on that basis to determine the spatial translation and rotation between the two views. Various techniques for the surface registration are proposed and they can be generally divided into the two groups: coarse and fine registration methods [1]. In the coarse registration the main goal is to compute an initial estimation of the rigid motion between the two views, while in the fine registration the goal is to obtain the most accurate solution by refining a known initial estimation of the solution.

The problem of finding the required rigid motion can be viewed as solving a six dimensional optimization problem (translation on x-, y- and z-axis, and rotation about x-, y- and z-axis). Genetic algorithms (GA) can be used to robustly solve this hard optimization problem and their advantage is that they are applicable both as coarse and fine registration methods [1].

In this work we propose a surface registration method assuming that the rotation is provided by an inertial sensor and the translation vector is still left to be found. We note that the registration method, based on the abovementioned assumption, has been successfully tested earlier [4]. However, in this work we specifically employ GA in that context. We justify our assumption by the fact that nowadays technologies have made quite affordable a large pallet of various inertial devices which reliably outputs data about the object orientation. For example, an inertial sensor can be explicitly used [2], or alternatively, there are also smart cameras which have an embedded on-board inertial sensor unit for the orientation detection in 3D space [3].

## II. Background

Genetic algorithm is a metaheuristic based on the concept of natural selection and evolution (Fig. 1.). GA represents each candidate solution as an individual that is defined by its parameters (i.e. its genetic material) and each candidate solution is qualitatively evaluated by the fitness function. Better solutions are more likely to reproduce, and the system of reproduction is defined by the crossover and mutation. The use of genetic algorithms has the advantage of avoiding the local minima which is a common problem in registration, especially when the initial motion is not provided. Another advantage of genetic algorithms is that they work well in the presence of noise and outliers given by the non-overlapping regions. The main problem of GA is the time required to converge, which makes it unsuitable for online applications.

Chow presented a dynamic genetic algorithm for surface registration where every candidate solution is described by a chromosome composed of the 6 parameters of the rigid motion that accurately aligns a pair of range images [5]. The chromosome is composed of the three components of the translation vector and the three angles of the rotation vector. In order to minimize the registration error, the median of Euclidean distances between corresponding points in two views is chosen as the fitness function. A similar method was proposed the same year by Silva [6]. The main advantage of this work is that the more robust fitness function is used and the initial guess is not required.

Crossover and mutation are basic genetic operators that are used to generate the new population of solutions from the previous population. In our work we use the crossover method that is accomplished by generating two random breaking points between six variables of the chromosome and mutation is defined as the addition or subtraction of a randomly generated parameter value to one of the variables of the chromosome. Probability for the mutation and the maximum value for which a variable can mutate dynamically change during the process of execution.

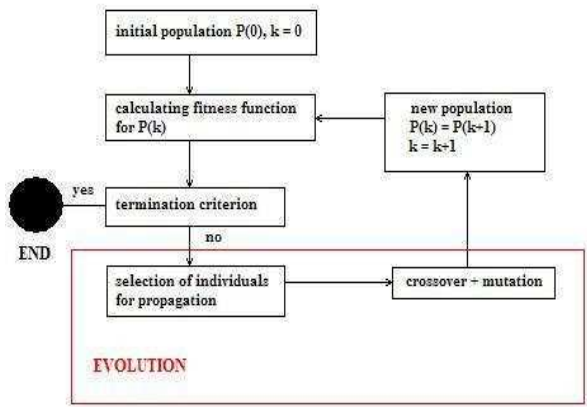

Fig. 1. Genetic algorithm

### III. PROPOSED METHOD

In this work we have used the same definition of the chromosome as was suggested by Chow in his paper [5]. The fitness function is defined as an overall mean of all the Euclidean distances between the corresponding points of the two clouds of 3D points for a given rigid motion. The objective of genetic algorithm is to find a candidate solution of the rigid motion that minimizes the overall mean distance between the corresponding points of the two views. Although Chow claims that it is better to use median of Euclidean distances between corresponding points instead of mean for evaluation of solutions, initial tests have proven that mean gives better practical results. This can be explained by the different use of GA, where Chow uses GA as a fine registration method, we use GA as both coarse and fine registration methods where in the absence of a good initial solution using median can cause algorithm to converge to a local minima more easily. Finally, we evaluate the proposed GA method assuming that the rotation is known in advance.

A. Implementation of genetic algorithm

We use two genetic algorithms in a sequence. First is used as the coarse registration method and produces an initial estimate of the solution, while other is used as the fine registration method (Fig. 2.).

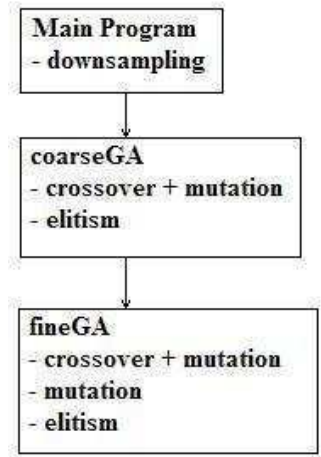

Fig. 2. Implementation of genetic algorithm

Population size and number of generations are two important factors of GA design. It is often cited in the literature that the population of 80 to 100 individuals is sufficient to solve more complex optimization problems [7]. Overpopulation may affect the ability of more optimal solutions to successfully reproduce (survival of the fittest) and thus prevent the convergence of algorithm towards the desired solution. If the population is too small then there's a greater danger that the algorithm will converge to some local extreme. After initial testing we have chosen parameters for GA as displayed in Table 1.

TABLE I. SIZE OF POPULATION

| GA | population | generations |
|---|---|---|
| coarseGA | 100 | 250 |
| fineGA | 50 | 250 |

CoarseGA function implements GA as a coarse registration method. Population is initialized by generating the initial correspondences in a way that the point that is closest to the center of mass from the first cloud of points is randomly matched with points in the second cloud of points. Translation vector for each such motion between views is determined. The angles of rotation vector are then initialized to randomly generated values where every solution is rotated only for one angle, or if we adopt limited space search, a priori known values of angles are inserted in the algorithm. In the coarseGA function elitism principle is implemented, transferring best two solutions directly to the next generation. The remaining generation is generated by the combined processes of crossover and mutation. Mutation probability for each gene is 16%. The algorithm is performed for fixed 250 generations.

FineGA inherits the last population of coarseGA. In the each iteration fineGA keeps the two best solutions, while 44 new individuals are generated by the processes of crossover and mutation and four new individuals are generated by the process of mutation only. Starting mutation probability is 20% and it increases for 5% every 25 iterations, while the maximum value for which the parameter can change decreases by a factor of 0.8. A rationale why we dynamically adjust the mutation parameters is that as the population converge to a good solution, smaller adjustments of genes are needed and therefore mutation becomes primal source for improvement of genetic material. That means that a mutation needs to happen more often, but at the same time it should produce less radical changes in genetic material.

Finding corresponding points in two clouds of 3D points is extremely time consuming operation, so to minimize a overall time of execution both downsampling of the 3D images and a KDtree search method are used. KDtree method for finding closest points in two clouds of points was suggested in earlier works [8] and has proven to be very fast compared to other methods.

## IV. RESULTS

Software was implemented in MATLAB development environment. 13 3D test images were obtained by partial reconstruction of a model doll using a structured light scanner. An algorithm was tested both in situation where no a priori information about rotational angles is available and where exact values of angles are known. In order to come up with the known ground true values we have manually marked 3-4 corresponding points on each pair of views. That allowed us to compute a rough estimate for the ground truth data which we have further refined using the well-known iterative closest point (ICP) algorithm. ICP is known to be very accurate given a good enough initial solution [9]. That was certainly the case in our experiments since we have manually chosen corresponding points very carefully. Therefore in the present context we regard such registration data as ground truth values. The basic algorithm (i.e. assuming no rotation data are known in advance) has proven successful for the registration of neighboring point clouds, but insufficiently accurate for a good registration between mutually distant views. An additional problem is a significant time that is needed for algorithm to produce the results. Improvement of the system, based on reduction of search space, has proven much more successful. Table 2. shows the average deviation of the results from the ground truth values when we search corresponding points in two more distant views (i.e. we compare views 3 and 1, 4 and 2, etc.). The first row represents the results when the rotation is not known (GA1) and the second row displays the results when the rotation is known (GA2). As algorithm finds three parameters more quickly than all six parameters, there is no need for GA to run through all 500 generations, which significantly decreases an average time of execution. What is more important, we get much better results. Tables 3. and 4. show results for views 4-2 and 13-11. The first row shows ground truth values, the second row shows values found by GA when the rotation is not known and the third row shows values when the rotation is known. The column named with % displays the percentage of the corresponding points between the two clouds of points. The results for the given views can also be seen on figures Fig. 3. – Fig. 6. Green and red images represent two different views, left part of the picture is before we applied the rigid motion between the views and the right part of the picture is after the rigid motion is applied and the images are rotated and translated for the values given by GA.

TABLE II. STANDRAD DEVIATION AND TIME OF EXECUTION

|  | x/mm | y/mm | z/mm | α/° | β/° | ψ/° | time/min |
|---|---|---|---|---|---|---|---|
| GA1 | 71 | 44 | 169 | 9 | 49 | 8 | 27 |
| GA2 | 22 | 19 | 15 | known | known | known | 8 |

TABLE III. RESULTS FOR VIEWS 4-2

|  | x | y | z | α | β | ψ | % |
|---|---|---|---|---|---|---|---|
| ground | 196 | 9 | -318 | 0 | 57 | 3 | 56 |
| GA1 | 177 | -77 | -99 | 5 | 20 | -1 | 28 |
| GA2 | 207 | 5 | -308 | known | known | known | 56 |

TABLE IV. RESULTS FOR VIEWS 13-11

|  | x | y | z | α | β | ψ | % |
|---|---|---|---|---|---|---|---|
| ground | 330 | 31 | -299 | 0 | 67 | 0 | 31 |
| GA1 | 118 | 62 | 141 | -1 | -57 | -8 | 17 |
| GA2 | 359 | -12 | -270 | known | known | known | 21 |

## V. CONCLUSION

The surface registration is an important step in the process of 3D object reconstruction and can be seen as an optimization problem with six degrees of freedom. The objective of the registration algorithms is to find the spatial translation and the rotation between the views so that two clouds of 3D points overlap with each other properly. Genetic algorithm has proven to be a suitable method for solving this particular optimization problem and can be used both as a coarse and fine registration method. The tests have shown that reducing the search space of the optimization problem from six to three parameters leads to better results and faster execution of the algorithm. The above is important because today's acquisition systems allow us to deploy the solutions where we can get the information about the rotational angles between the different views. Therefore, our next research objective will be a 3D reconstruction system design where rotation data are readily available and the surface registration reduces to computing the translation part only.

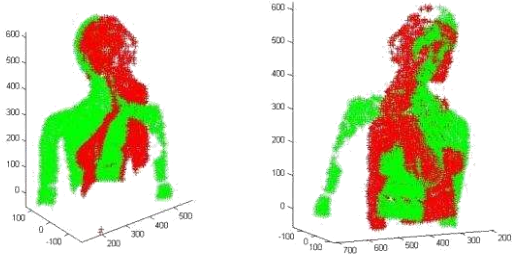

Fig. 3. Views 4-2, unknown rotation

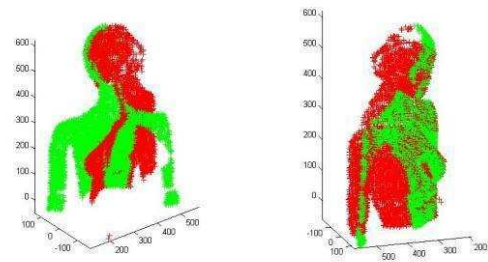

Fig. 4. Views 4-2, assuming known rotation in advance

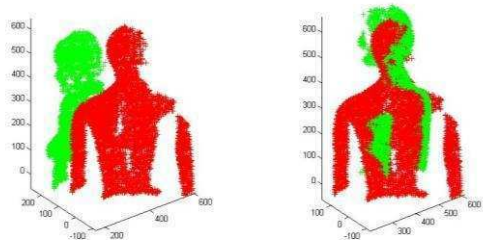

Fig. 5. Views 13-11, unknown rotation

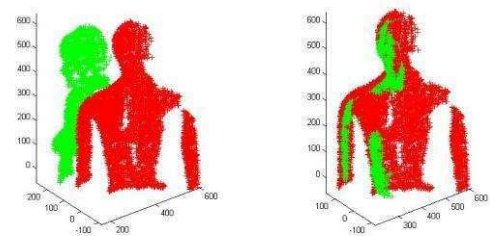

Fig. 6. Views 13-11, assuming known rotation in advance